# Comprehensive Pathological Image Segmentation via Teacher Aggregation for Tumor Microenvironment Analysis

Daisuke Komura[1], Maki Takao[1,2], Mieko Ochi[1], Takumi Onoyama[1,3], Hiroto Katoh[1], Hiroyuki Abe[4], Hiroyuki Sano[5], Teppei Konishi[5], Toshio Kumasaka[6], Tomoyuki Yokose[7], Yohei Miyagi[8], Tetsuo Ushiku[4], & Shumpei Ishikawa[1,9,*]

[1]Department of Preventive Medicine, Graduate School of Medicine, The University of Tokyo, 7-3-1 Hongo, Bunkyo-ku, 113-0033, Tokyo, Japan

[2]Comprehensive Reproductive Medicine, Graduate School of Medicine and Dental Sciences, Institute of Science Tokyo, 1-5-45, Yushima, Bunkyo-ku, 113-8510, Tokyo, Japan

[3]Division of Gastroenterology and Nephrology, Department of Multidisciplinary Internal Medicine, School of Medicine, Faculty of Medicine, Tottori University, 36-1 Nishicho, Yonago, 683-8504, Tottori, Japan

[4]Department of Pathology, Graduate School of Medicine, The University of Tokyo, 7-3-1, Hongo, Bunkyo-ku, 113-0033, Tokyo, Japan

[5]Biomy Inc., 3-8-3, Nihonbashi-honcho, Chuo-ku, 103-0023, Tokyo, Japan

[6]Department of Pathology, Japanese Red Cross Medical Center, 4-1-22, Hiro-o, Shibuya-ku, 150-8935, Tokyo, Japan

[7]Department of Pathology, Kanagawa Cancer Center, 2-3-2, Nakao, Asahi-ku, Yokohama, 241-8515, Kanagawa, Japan

[8]Kanagawa Cancer Center Research Institute, 2-3-2, Nakao, Asahi-ku, Yokohama, 241-8515, Kanagawa, Japan

[9]Division of Pathology, National Cancer Center Exploratory Oncology Research & Clinical Trial Center, 6-5-1, Kashiwanoha, Kashiwa, 277-8577, Chiba, Japan

**Correspondence:** ishum-prm@m.u-tokyo.ac.jp

## Competing Interests

The authors declare no competing interests.

**Abstract**

The tumor microenvironment (TME) plays a crucial role in cancer progression and treatment response, yet current methods for its comprehensive analysis in H&E-stained tissue slides face significant limitations in the diversity of tissue cell types and accuracy. Here, we present PAGET (Pathological image segmentation via AGgrEgated Teachers), a new knowledge distillation approach that integrates multiple segmentation models while considering the hierarchical nature of cell types in the TME. By leveraging a unique dataset created through immunohistochemical restaining techniques and existing segmentation models, PAGET enables simultaneous identification and classification of 14 key TME components. We demonstrate PAGET's ability to perform rapid, comprehensive TME segmentation across various tissue types and medical institutions, advancing the quantitative analysis of tumor microenvironments. This method represents a significant step forward in enhancing our understanding of cancer biology and supporting precise clinical decision-making from large-scale histopathology images.

**Introduction**

In 1889, Stephen Paget proposed the "seed and soil" hypothesis, suggesting that cancer cells (the "seed") can only grow in a favorable environment (the "soil"). This pioneering concept laid the foundation for our current understanding of the tumor microenvironment (TME). Today, we recognize that the TME plays a crucial role in cancer progression, treatment response, and patient prognosis[1]. This dynamic ecosystem comprises various cell types and extracellular matrix components. Recent research has emphasized the significance of characterizing the TME, particularly its immune cell composition and spatial distribution, as potential prognostic and predictive biomarkers[2–4]. Furthermore, other stromal cells, such as cancer-associated fibroblasts (CAFs) and tumor-associated endothelium (TAE), have also emerged as key players in modulating the TME, contributing to tumor progression through diverse mechanisms[5,6].

The spatial relationships between these various cell types can significantly influence their functions and interactions. For example, the proximity of lymphocytes to tumor cells may indicate an active immune response. Similarly, the distribution of other cell types like plasma cells, neutrophils, or eosinophils in relation to tumor cells and stromal elements can provide important clues about the nature of the immune response and the overall

TME dynamics.

Therefore, there is a growing need for sophisticated tools to accurately characterize and quantify the diverse cellular components within the TME in hematoxylin and eosin (H&E) stained tissue slides[7–9], which are abundantly generated and routinely used in pathological diagnosis. Traditional histopathological assessment of TME faces the challenge of comprehensively analyzing large tissue sections containing hundreds of thousands of cells. Recent advancements in digital pathology and artificial intelligence have revolutionized histopathology, enabling sophisticated analysis of tissue architecture and cellular composition. Various segmentation models have demonstrated the ability to identify and classify various cell types within tumor tissues[10,11]. However, current segmentation models have several limitations. They rely heavily on morphology-based annotations by pathologists, which can introduce bias and inaccuracies, particularly for cell types that are difficult to distinguish visually. Additionally, these models are limited in the range of tissue structures and cell types they can reliably identify, often focusing on a relatively small subset of the diverse cellular components that make up the tumor microenvironment (TME). As a result, the current segmentation approaches may fail to capture the full complexity and heterogeneity of the TME.

To address this challenge, we previously presented an annotation technique using immunohistochemical restaining[12]. This method allows for annotations based on the expression of proteins that define cellular identity, rather than relying solely on morphological features. While this approach improved the accuracy of segmentation and enabled the recognition of a broader range of tissue structures and cell types, it does not allow for the annotation of all cell types within a single tissue image, necessitating the development of separate models for each cell type. This limitation precludes the practical application of this method to large-scale whole slide image (WSI) analysis due to the increased computational time and resources required.

To overcome the limitation, here we present PAGET (Pathological image segmentation via AGgrEgated Teachers), a novel approach that integrates multiple segmentation models while considering the hierarchical nature of cell types in the TME. Models trained with PAGET offer simultaneous identification and classification of a broader range of cell and tissue types, enabling more detailed and complex histological analyses. Our method leverages a unique dataset created using the immunohistochemical restaining technique as well as existing segmentation models, encompassing 14 key components of the TME. By offering a single, efficient model capable of accurate multi-class segmentation, PAGET represents a significant advancement in the quantitative analysis of the TME.

We demonstrate PAGET model's ability to perform rapid, comprehensive segmentation of the TME across various tissue types and medical institutions. This approach aims to enhance the efficiency and depth of tumor microenvironment analysis, potentially advancing our understanding of cancer biology and supporting more precise clinical decision-making.

**Materials and methods**

*Training dataset*

A subset of the SegPath dataset, specifically the H&E stained image data generated for CD3/CD20 and CD45 immunohistochemistry, were used to train the PAGET model. All histopathological specimens were obtained from patients diagnosed between 1955 and 2018 who had undergone surgery at the University of Tokyo Hospital. Detailed information is found in the previous paper[12]. In brief, tissue-containing regions were identified from Tissue Microarrays (TMAs), from which patches were extracted. The dataset comprised 22 and 18 TMA slides for CD3/CD20 and CD45, respectively. Tissue types in the dataset are shown in Table 1. The final dataset consisted of 59,443 images for training and 3,133 images for validation. While the original pixel size was 984×984 pixels at 40× magnification, the images were resized to 492×492 pixels for training.

**Table 1 Number of training images in each tumor type.**

| Cancer Type | Image Count |
| --- | --- |
| Endometrial cancer | 3347 |
| Breast cancer | 3264 |
| Bladder cancer | 2884 |
| Urothelial tumor | 2873 |
| Prostate cancer | 2790 |
| Kidney tumor | 2783 |
| Gastric cancer | 2679 |
| Extrahepatic bile duct cancer | 2517 |
| Colorectal cancer | 2290 |
| Triple-negative breast cancer | 2046 |
| Esophagogastric junction cancer | 2035 |
| Gastric cancer lymph node metastasis | 1911 |
| Lung squamous cell carcinoma | 1852 |
| Benign breast lesion | 1831 |
| Pancreatic cancer | 1785 |
| Hypopharyngeal and laryngeal cancer | 1747 |
| Hepatocellular carcinoma | 1723 |
| Cervical squamous cell carcinoma | 1709 |
| Pancreatic neuroendocrine tumor | 1675 |
| Esophagogastric junction cancer | 1666 |
| Pancreatic cancer | 1643 |
| Ependymoma | 1563 |
| Background liver | 1539 |
| Early colorectal cancer | 1528 |
| Colorectal cancer | 1406 |
| Pancreatic IPMN + neuroendocrine tumor | 1332 |
| Extrahepatic bile duct cancer | 1302 |
| Early gastric cancer | 1265 |
| Liver cancer | 1220 |
| Thymoma | 667 |
| Ovarian mucinous cystic neoplasm | 575 |
| total | 59447 |

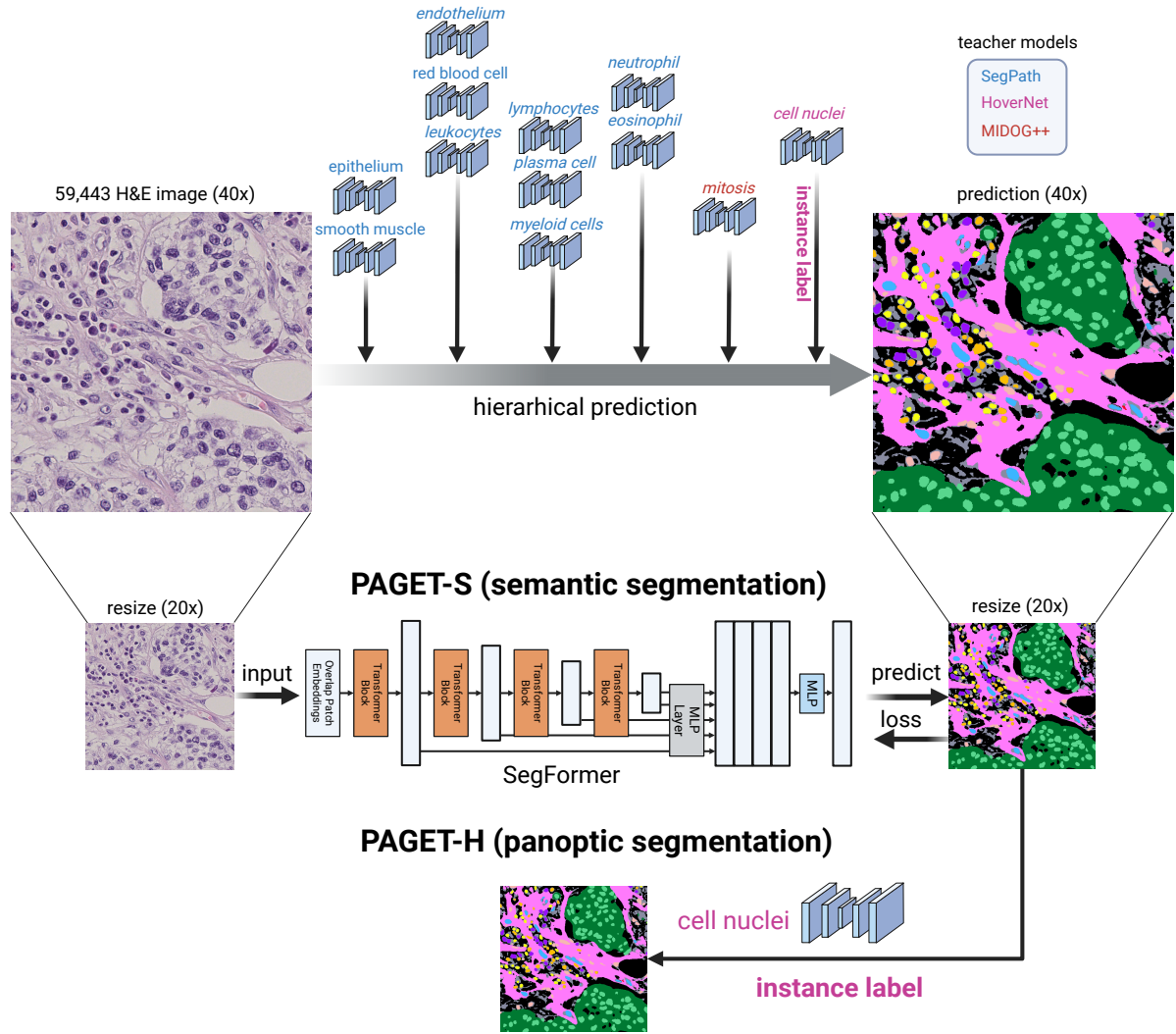

**Fig. 1 Overview of PAGET.** a. Flow chart of PAGET model training. The process begins with a 40× H&E image and the prediction by specialized models for different cell types. These predictions are hierarchically aggregated to create teacher data. The PAGET model is then trained on resized 20× images to predict multi-class semantic segmentation, with loss calculated against the teacher data. b. Two PAGET model variants: PAGET-S performs semantic segmentation, while PAGET-H conducts panoptic segmentation by integrating semantic segmentation results with nucleus instance segmentation by HoverNet.

*Cell hierarchy-aware aggregated distillation*

We developed a cell hierarchy-aware aggregated distillation approach, utilizing outputs from various teacher models for training. The teacher models comprised semantic segmentation models trained on an extended version of SegPath, a publicly available mitosis detection model trained on MIDOG++[13], and the HoverNet model[10] which was trained on the PanNuke dataset[14] for nucleus instance segmentation and classification. Each teacher model, except for HoverNet, is specialized in segmenting a single tissue or cell type.

The training details for the SegPath models, which include epithelium, smooth muscle tissue, red blood cells, endothelial cells, leukocytes, lymphocytes, myeloid cells, and plasma cells, were previously described in our earlier publication[12]. To identify additional key cellular components in the TME, we developed semantic segmentation models for neutrophils and eosinophils in an extended SegPath dataset using anti-Myeloperoxidase and anti-Eosinophil Cationic Protein (ECP) antibodies, respectively. Samples of the mask for neutrophils and eosinophils are shown in Supplementary Fig 1.

For the neutrophil model, we employed an EfficientNet-B1[15] architecture with noisy student pretraining[16], implemented in a U-Net structure[17]. We utilized the Dice loss function, achieving a validation Dice score of 0.411. The eosinophil model was based on a ResNet34 architecture pretrained on ImageNet, incorporated into a DeepLabV3Plus framework[18]. This model also used Dice loss, resulting in a validation Dice score of 0.299. The SegPath models were trained on a server with a single NVIDIA A100 40GB GPU, two Intel Xeon Platinum 8360Y 2.4GHz processors, and 512GB of memory.

MIDOG++[13] and HoverNet[10] models were used to detect mitotic cells and fibroblasts, which are not included in the SegPath dataset, respectively. HoverNet model was also used as nucleus instance segmentation to improve nucleus detection accuracy in the PAGET model.

We developed an integrated segmentation approach that considers the hierarchical organization of the TME components. Our method processes the image in two main stages: tissue-level segmentation and cellular classification.

The tissue-level segmentation begins with nuclear detection using HoverNet and background identification using Otsu's thresholding on Gaussian-smoothed H&E images. We then apply dedicated models for smooth muscle and epithelial tissues, classifying regions based on their logit values. Red blood cells are detected and overlaid, with remaining unclassified regions designated as stroma.

For cellular classification, we employ a four-level hierarchical scheme for each detected nucleus:

1. Tissue context: smooth muscle and epithelial tissues

2. Major cell categories: leukocytes, endothelial cells, and red blood cells

3. Leukocyte subtypes: lymphocytes, plasma cells, and myeloid cells

4. Granulocyte subtypes: eosinophils and neutrophils

At each level, classification is determined by the highest positive logit value. Classifications at lower levels can override higher-level assignments if their logit values are positive, allowing for more specific cell type identification. The final class for each nucleus is determined by majority voting of pixels within the nuclear region. Additionally, nuclei within epithelial tissue that remain unclassified are labeled as epithelial cells, while nuclei in stromal regions predicted as connective tissue by HoverNet are designated as fibroblasts.

To incorporate mitosis detection capabilities into our segmentation framework, we adapted the MIDOG++ model's output through a multi-step transformation process. First, we generate circular regions of interest (ROIs) with a 30-pixel radius around each potential mitotic figure detected by MIDOG++. These ROIs undergo filtering to remove false positives from carbon dust particles by excluding regions where the RGB sum is $\leq 40$. We then apply Otsu's thresholding to the remaining ROIs and identify contours, retaining only those with an area of $\geq 3$ pixels. The convex hull is computed for each valid contour, and regions overlapping with epithelial tissue are classified as mitotic. Each mitotic region is assigned a unique identifier. In the final step, any nucleus that intersects with a mitotic region is reclassified as a mitotic figure, superseding previous classifications. The complete algorithm is detailed in Figure 2. The number of nuclei and pixel count of tissue area in the training dataset are presented in Tables 2 and 3.

```
// Algorithm: Generate semantic segmentation masks from multiple teacher models

// Input:
  - H&E_image: Hematoxylin and Eosin stained histology image
  - HoverNet: Pre-trained model for nuclear detection
  - TissueModels: Pre-trained models for tissue classification
  - CellModels: Pre-trained models for cell type classification
  - MIDOG_PlusPlus: Pre-trained model for tsis detection

// Output:
  - SegmentationMask: Unified mask with classified TME components

// Functions:
function UnifiedTMESegmentation(H&E_image):
    // 1. Nuclear and background segmentation
    NuclearMask, CellTypesHoverNet = HoverNet(H&E_image)
    BackgroundMask = OtsuThreshold(GaussianSmooth(H&E_image))

    // 2. Tissue-level segmentation
    TissueMask = ApplyTissueModels(H&E_image, TissueModels)
    TissueMask = OverlayRedBloodCells(TissueMask, H&E_image)
    StromaMask = NOT(OR(BackgroundMask, TissueMask))

    // 3. Cellular-level classification
    for each Nucleus, CellTypeHovernet in NuclearMask, CellTypesHoverNet:
        CellType = HierarchicalCellClassification(Nucleus, CellModels)
        if CellType == 'epithelial' and NoOtherClassification(Nucleus):
            CellType = 'epithelial_cell'
        elif CellTypeHovernet == 'connective' and NoOtherClassification(Nucleus):
            CellType = 'fibroblast'
        AssignCellType(Nucleus, CellType)

    // 4. Mitosis detection
    MitosisCandidates = MIDOG_PlusPlus(H&E_image)
    MitosisMask = DetectMitosis(MitosisCandidates, H&E_image, TissueMask)

    // 5. Final classification
    for each Nucleus in NuclearMask:
        if Overlaps(Nucleus, MitosisMask):
            AssignCellType(Nucleus, 'mitotic_figure')

    // 6. Combine all masks
    SegmentationMask = CombineMasks(BackgroundMask, TissueMask, StromaMask,
                                    NuclearMask, MitosisMask)
    return SegmentationMask

function HierarchicalCellClassification(Nucleus, CellModels):
    Hierarchies = [
        ['smooth_muscle', 'epithelial'],
        ['leukocyte', 'endothelial', 'red_blood_cell'],
        ['lymphocyte', 'plasma_cell', 'myeloid_cell'],
        ['eosinophil', 'neutrophil']
    ]
    for Hierarchy in Hierarchies:
        ClassProbabilities = ApplyCellModels(Nucleus, CellModels[Hierarchy])
        if max(ClassProbabilities) > 0:
            return argmax(ClassProbabilities)
    return 'undefined'

function DetectMitosis(MitosisCandidates, H&E_image, TissueMask):
    MitosisMask = EmptyMask()
    for Candidate in MitosisCandidates:
        ROI = CircularROI(Candidate, radius=30)
        if not ExcessiveDarkPixels(ROI) and OverlapsEpithelial(ROI, TissueMask):
            ContourMask = ProcessROI(ROI)
            MitosisMask = AddToMask(MitosisMask, ContourMask)
    return MitosisMask

// Main execution
SegmentationResult = UnifiedTMESegmentation(InputH&E_image)
```

**Fig. 2 Pseudocode for generating semantic segmentation masks from multiple teacher models.**

**Table 2 Number of nuclei in training data**

| nucleus type | number of nuclei |
|---|---:|
| epithelial cell | 8,388,735 |
| leukocyte | 1,285,388 |
| lymphocyte | 1,205,822 |
| plasma cell | 357,362 |
| myeloid cell | 461,106 |
| eosinophil | 36,853 |
| neutrophil | 291,370 |
| endothelial cell | 484,132 |
| fibroblast | 2,914,845 |
| mitotic cell | 2,726 |
| total | 15,428,339 |

**Table 3 Pixel count of tissue in training data**

| tissue type | pixels |
|---|---:|
| Epithelium | 4,031,758,228 |
| Stroma | 2,464,656,297 |
| Smooth muscle | 2,081,948,879 |
| Red blood cell | 130,403,380 |
| total | 8,708,766,784 |

*Student model training*

For the training of PAGET student model, we employed a Segformer architecture[19] featuring a mixed transformer (MiT-B5) encoder pretrained on ImageNet-1K, and a decoder with channel dimensions of 64, 128, 320, and 512. The preprocessing stage involved cropping input images to 384×384 pixels, in accordance with our data augmentation strategy, and using a stride of 320×320.

We utilized the AdamW optimizer with default settings, a learning rate of 0.00006, betas of (0.9, 0.999), and a weight decay of 0.01. Our learning rate schedule combined a linear warmup phase from 0 to 1500 iterations, starting with a factor of 1e-6, followed by a polynomial decay phase from 1500 to 48,0000 iterations, with a minimum learning rate of 0 and a power of 1.0.

CrossEntropyLoss was used as a loss function. The training process used a batch size of 4 and 8 worker threads for data loading. The data augmentation pipeline included random resizing (0.85 to 1.15), random cropping with a category max ratio of 0.75, random horizontal and vertical flipping with a probability of 0.5, random blurring, random gamma adjustment, and photometric distortions (brightness, contrast, and hue-saturation adjustments, with a hue delta of 36).

The models were trained on a server with eight NVIDIA H100 80GB GPUs, two Intel Xeon Platinum 8480+ 2.0GHz processors, and 1,024GB of memory using PyTorch 2.2 and MMSegmentation 1.2 with Python 3.8 on CUDA 12.1.

*Inference of PAGET models*

To facilitate downstream analysis and interpretation, pixels initially classified as leukocytes were reassigned to the non-leukocyte blood cell class with the highest logit value. For PAGET-H, we further refined the classification using a HoVer-Net model to detect individual nuclei. Each detected nucleus was assigned to the class with the highest sum of logit values across its pixels. Regions not identified as nuclei were classified on a pixel-by-pixel basis, assigning each pixel to the non-nucleus class with the highest logit value.

*Evaluation of PAGET models*

The evaluation of PAGET was conducted using two publicly available datasets, PanopTILs[20] (manual annotation), Lizard[21], and an original external cohort (KCCRC). The PanopTILs dataset consisted of breast cancer samples, the Lizard dataset contained five subsets (PanNuke, DigestPath, GlaS, CoNSeP, CRAG) from colorectal cancer samples, and the original cohort included cases of goblet cell carcinoid and pseudomembranous enterocolitis from Japanese Red Cross Medical Center, as well as colorectal cancer from Kanagawa Cancer Center. For the Lizard dataset, since HoVerNet trained on the PanNuke dataset was used for the label creation in the PAGET training, the PanNuke subset was excluded from the evaluation.

The comparative models included HD-Yolo, HoverNet trained on the PanNuke or MoNuSAC datast, and Cerberus[22]. Cerberus and PAGET-S, PAGET-H correspond to 20× image resolution, while HD-Yolo and the HoverNet models only support 40× image resolution. Since the Lizard dataset only had 20× images, the 40×-compatible models were upsampled using the BICUBIC method to match the image resolution. Since Cerberus was trained on Lizard dataset, we excluded Cerberus from the evaluation in Lizard dataset.

Due to the varying class definitions across the different models, the classes in each dataset were mapped to the corresponding classes of the respective models to ensure a biologically meaningful evaluation (Supplementary Table 1).

We employed two metrics:

1. Dice Score: Used for assessing tissue segmentation accuracy. The Dice score is defined as: Dice = (2 × |X ∩ Y|) / (|X| + |Y|) where X and Y are the predicted and ground truth segmentation masks, respectively. This score ranges from 0 to 1, with 1 indicating perfect overlap.

2. Matthew's Correlation Coefficient (MCC): Employed for evaluating nucleus instance classification accuracy. The MCC is calculated for each nucleus in the ground truth of each dataset. It is defined as: MCC = (TP * TN - FP * FN) / sqrt((TP + FP) * (TP + FN) * (TN + FP) * (TN + FN)) where TP, TN, FP, and FN are true positives, true negatives, false positives, and false negatives, respectively. The MCC ranges from -1 to 1, with 1 representing a perfect prediction, 0 indicating no better than random prediction, and -1 representing total disagreement between prediction and observation.

The MCC was chosen for its advantage as a balanced measure[23]: It provides a balanced evaluation of the quality of binary (and by extension, multi-class) classifications, even when classes are of very different sizes. The MCC evaluation was performed for the class with the largest pixel coverage within the ground truth instances.

*Approximated Prediction of Cell Count in PAGET-S*

While PAGET-S is primarily a semantic segmentation model, we explored a method, which utilized the pixel count of the respective cell nuclei. This approach assumes that the number of pixels corresponding to a particular cell type is proportional to the number of cells present.

*Correlation between tumor microenvironment features and somatic mutations of uterine corpus endometrial carcinoma*

We analyzed 505 diagnostic H&E slides from 445 cases of uterine corpus endometrial carcinoma (UCEC) from The Cancer Genome Atlas (TCGA) were processed using PAGET-S. Given the minimal presence of normal epithelial cells in the UCEC cases, all detected epithelial cells were classified as tumor cells for TME metric calculations Cell counts were determined by the number of connected regions identified in the predicted segmentation results. Two categories of metrics were calculated to characterize the tumor microenvironment: 1) the ratio of each cell type count to tumor cell counts across the entire slide, and 2) the ratio of specific cell type densities (including fibroblasts, endothelial cells, lymphocytes, plasma cells, myeloid cells, neutrophils, eosinophils, and all leukocytes) within 50 μm of the tumor margin relative to the total tumor cell count. For cases with multiple whole slide images, mean values were used.

Clinical and genomic information for the UCEC cases was obtained from the cBioPortal database[24]. The dataset encompassed 507 genes, with 310 classified as driver mutations. Our analysis focused on 51 driver genes, comprising 46 genes that were mutated in more than 5% of cases, plus five additional genes of interest: MLH1, MSH2, PMS2, BRCA1, and SMARCA4 (Supplementary Table 2).

**Results and Discussion**

*Segmentation model development with cell hierarchy-aware aggregated distillation*

We developed PAGET (Pathological image segmentation via AGgrEgated Teachers), to train a multi-class segmentation model that incorporates cell hierarchy-aware aggregated distillation. Figure 1 provides an overview of the PAGET and its training process.

Our approach utilizes a comprehensive set of 14 different tissue or cell types present in the tumor microenvironment, including epithelium (tissue and cell), stroma, smooth muscle tissue, endothelium, red blood cells, leukocytes, lymphocytes, neutrophils, eosinophils, plasma cells, myeloid cells, fibroblasts, and mitotic cells. The training process begins with a high-resolution H&E image (40× magnification) as input. This image is then processed through a hierarchical prediction pipeline, where individual models for each cell type make predictions. These predictions are subsequently aggregated using the hierarchical structure of cell types to produce a final segmentation output. This integrated approach enables comprehensive segmentation and classification of various tissue and cell types, providing a detailed characterization of the tumor microenvironment from H&E-stained tissue slides. Our method leverages the strengths of multiple models, including HoverNet for nuclear detection, specialized models for tissue classification, and a hierarchical approach for cellular identification. By combining these elements, we achieve a more accurate and comprehensive analysis of the TME, capturing the complex interplay of different cell types and structures within the tumor landscape.

The PAGET model is trained using this aggregated output as teacher data. The model's ability to take a resized 20× version of the original image as input offers two key benefits - it aligns with the common use of 20× magnification in clinical scan images, making the model well-suited for real-world application, and it enables faster processing compared to higher 40× resolutions, which is advantageous for practical deployment in time-sensitive clinical environments. The loss between the prediction and the hard label predicted by teacher models is used to update the model parameters.

We developed two variants of the PAGET model (Fig. 1b):

1. PAGET-S: This variant performs semantic segmentation, providing a pixel-wise classification of the image into different cell types.

2. PAGET-H: This variant performs panoptic segmentation by integrating the results of semantic segmentation with nucleus instance segmentation, allowing for both classification and individual cell identification at the cost of increased processing time.

Fig.3 shows some samples of teacher labels and its prediction of PAGET models. Table 4 presents the Intersection over Union (IoU) values for PAGET-S and PAGET-H on the internal test dataset.

**Table 4 Internal test performance of PAGET-S**

| Class | Abbreviation | PAGET-S IoU | PAGET-H IoU |
|---|---|---|---|
| background | bg | 0.847 | 0.848 |
| stroma | str | 0.709 | 0.715 |
| smooth muscle | sm | 0.822 | 0.814 |
| epithelial tissue | epi | 0.772 | 0.809 |
| leukocyte | leu | 0.504 | 0.569 |
| endothelial cell | endo | 0.585 | 0.532 |
| red blood cell | rbc | 0.805 | 0.783 |
| lymphocyte | lym | 0.646 | 0.753 |
| plasma cell | pls | 0.556 | 0.612 |
| myeloid cell | mye | 0.399 | 0.450 |
| eosinophil | eos | 0.440 | 0.525 |
| neutrophil | neu | 0.538 | 0.605 |
| epithelial cell nucleus | epi_n | 0.760 | 0.853 |
| fibroblast | fib | 0.613 | 0.649 |
| mitotic cell | mit | 0.302 | 0.382 |

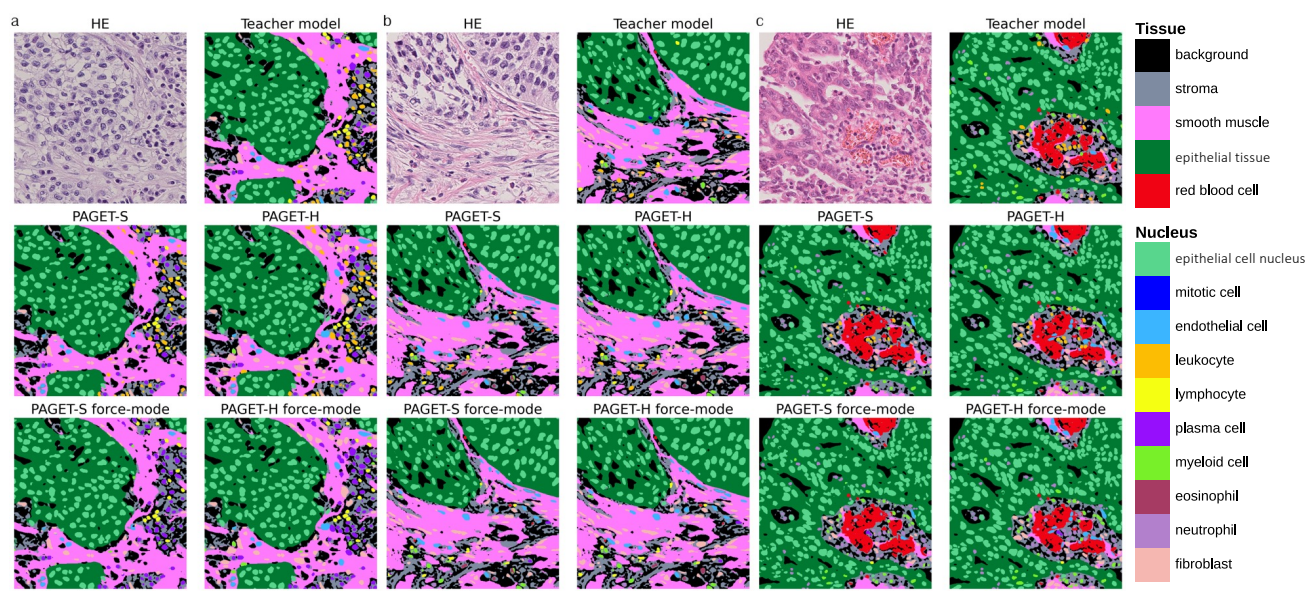

**Fig. 3 Representative H&E images with corresponding predictions by teacher models, PAGET-S, and PAGET-H.** The "Force-mode" demonstrates predictions where leukocytes are reclassified as detailed white blood cell types.

*PAGETs demonstrates improved segmentation performance in external dataset*

We evaluated PAGET-S and PAGET-H against three existing methods for cell segmentation and classification using diverse datasets and teacher model predictions. The evaluation included epithelial tissues, stromal tissues, blood (red blood cells), epithelium nuclei, connective tissue cells, leukocytes, lymphocytes, plasma cells, myeloid cells, eosinophils, and neutrophils.

To compare models with different class definitions, we applied a hierarchical classification approach. Leukocytes were grouped into lymphocytes and other white blood cells, while connective tissue cells included both fibroblasts and endothelial cells.

Table 4 and 5 compare epithelium and nucleus detection performance among different models on breast cancer using the Dice coefficient and Matthew's correlation coefficient as evaluation metrics, respectively. PAGET-H and PAGET-S achieved superior scores for epithelium nuclei detection compared to HD-Yolo and HoverNet, while surpassing Cerberus in epithelial tissue segmentation. This improved performance extended to challenging cell types such as plasma cells, myeloid cells, and eosinophils, where other models failed to provide reliable results or did not include the target cells (denoted by 'nan' in the table). Figure 4 demonstrates segmentation results from various models in the PanopTILs dataset. HoverNet and Cerberus tend to misidentify cells outside tumor tissue as tumor cells, which can be problematic when analyzing the tumor microenvironment (TME), especially near tumors. PAGET models, using Pan-CK as a teacher, can accurately identify epithelial tissue regions. This high accuracy in epithelial tissue segmentation by PAGET is consistent across datasets. Additionally, PAGET can detect mitotic cells, as shown in the upper left of Fig. 4a. The image in Fig. 4c contains many plasma cells, which PAGET correctly captures. Other models either fail to capture lymphocytes and plasma cells or, like Cerberus, misidentify them as epithelial cells.

The enhanced segmentation and classification capabilities of PAGET models were not limited to a single tissue type or dataset (Tables 6 and 7). Performance improvements were consistently observed across different tissue contexts and multi-cancer cohorts, as evidenced by results in the CoNSeP, CRAG, DigestPath, and KCCRC datasets. These diverse datasets represent various tissue types and pathological conditions, underscoring the robustness and generalizability of our approach.

Interestingly, both PAGET-S and PAGET-H outperformed their teacher models in almost all categories. This is likely due to these models being "noisy students,"[16] trained with heavier data augmentation, including color jitters.

Both PAGET-S and PAGET-H exhibited comparable performance levels, with subtle variations depending on the specific cell type and dataset. This consistency across model variants suggests that the core principles of our cell hierarchy-aware aggregated distillation approach effectively capture the complexities of cellular morphology and organization in various pathological contexts.

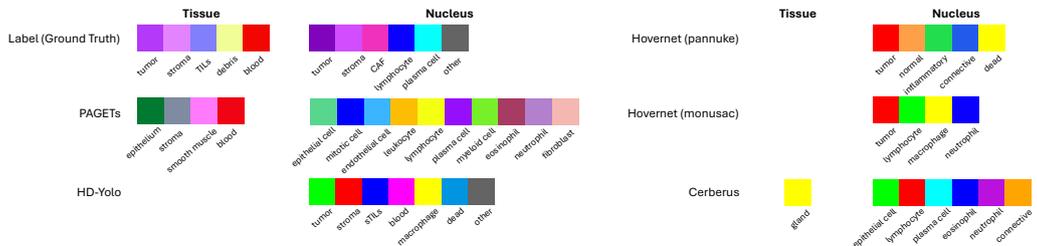
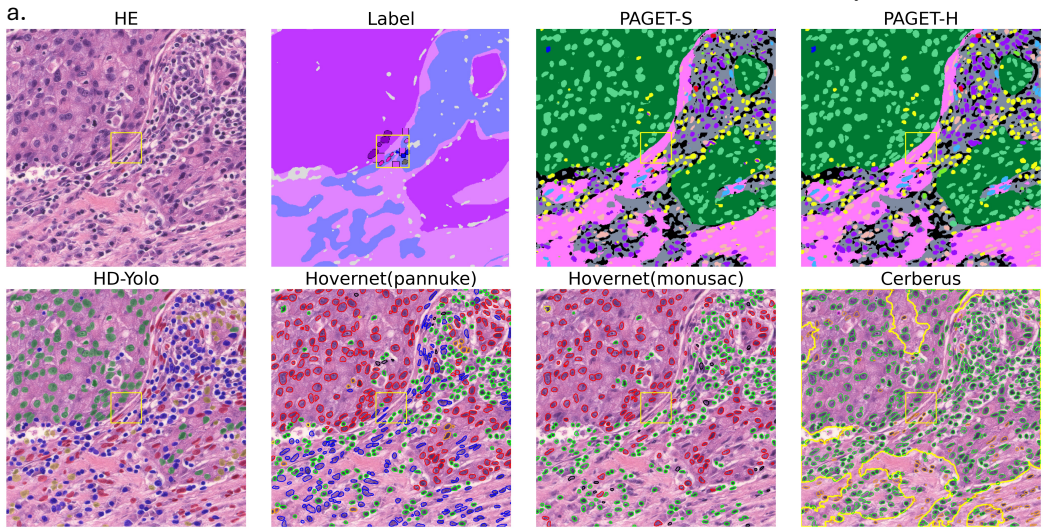
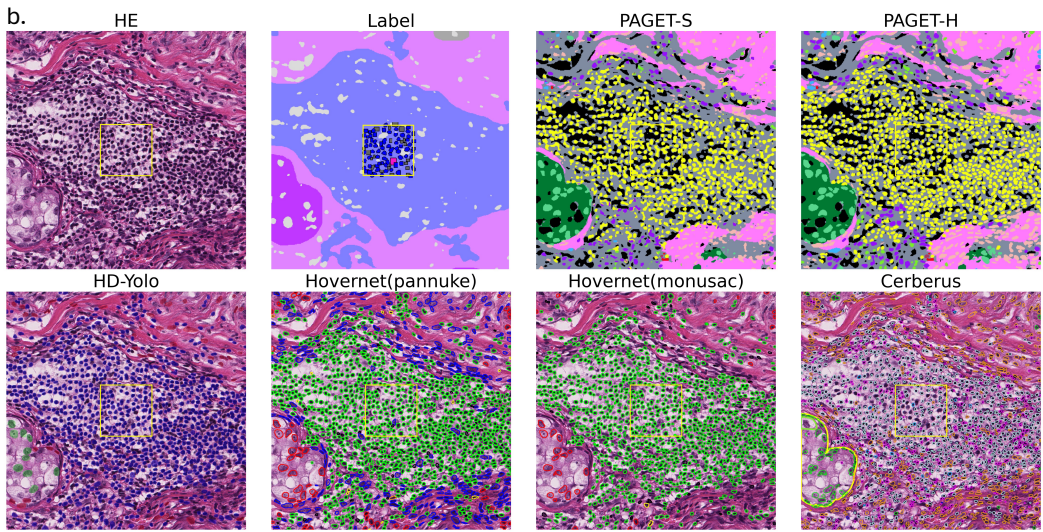
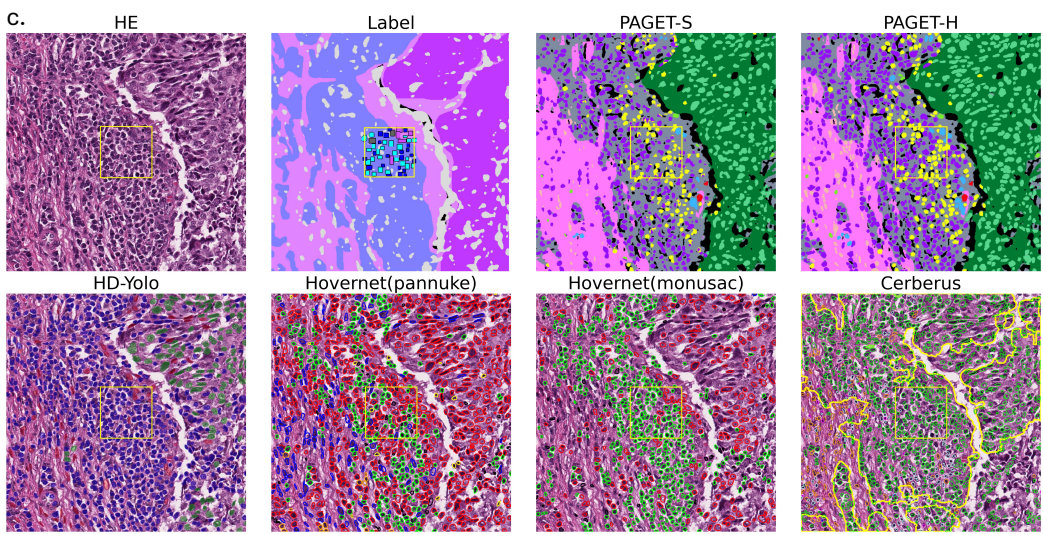

**Fig. 4 Samples of segmentation results in PanopTILs.** Representative H&E-stained tissue images are shown alongside their corresponding PanopTILs ground truth labels, segmentation results from the PAGET methods, and results from other segmentation models.

**Table 4 Dice index in the PanopTILs dataset.** The best performance in each column is bold, and the second best is in underlined. *This model was trained on a colon cancer dataset.

| Model | epithelial tissue | stroma | blood |
|---|---|---|---|
| **PAGET-S** | 0.868 | 0.694 | 0.443 |
| **PAGET-H** | **0.868** | **0.696** | 0.450 |
| *Teachers* | 0.797 | n/a | **0.465** |
| Cerberus | 0.800 | n/a | n/a |

**Table 5 MCC values in the PanopTILs dataset.** The best performance in each column is bold, and the second best is in underlined.

| Model | epithelial cell | connective tissue cell | leukocyte | lymphocyte | plasma cell |
|---|---|---|---|---|---|
| **PAGET-S** | 0.657 | 0.378 | 0.562 | 0.406 | **0.228** |
| **PAGET-H** | **0.663** | **0.415** | **0.579** | **0.424** | 0.226 |
| *Teachers* | 0.435 | 0.295 | 0.345 | 0.209 | 0.176 |
| HD-Yolo (breast) | 0.512 | 0.389 | 0.449 | 0.356 | n/a |
| Hovernet (pannuke) | 0.461 | 0.316 | 0.400 | n/a | n/a |
| Hovernet (monusac) | 0.438 | n/a | 0.424 | 0.368 | n/a |
| Cerberus | 0.437 | 0.210 | 0.367 | 0.311 | 0.121 |

**Table 6 MCC values in each subset in the Lizard dataset.** Bold and italic values indicate the models with the highest and second-highest performance, respectively, for each cell type.

| Subset | Model | epithelial cell | connective tissue cell | leukocyte | lymphocyte | plasma cell | eosinophil | neutrophil |
|---|---|---|---|---|---|---|---|---|
| **DigestPath** | | | | | | | | |
| | **PAGET-S** | **0.784** | 0.402 | 0.481 | 0.279 | **0.313** | 0.242 | **0.123** |
| | **PAGET-H** | 0.777 | 0.423 | 0.550 | 0.320 | 0.302 | **0.255** | 0.099 |
| | *Teachers* | 0.539 | 0.376 | 0.480 | 0.272 | 0.187 | 0.000 | 0.016 |
| | Hovernet (pannuke) | 0.476 | 0.402 | 0.482 | n/a | n/a | n/a | n/a |
| | Hovernet (monusac) | 0.236 | n/a | 0.446 | **0.449** | n/a | n/a | 0.048 |
| | HD-Yolo (lung) | 0.295 | 0.098 | 0.190 | 0.184 | n/a | n/a | n/a |
| | HD-Yolo (breast) | 0.424 | 0.281 | 0.354 | 0.290 | n/a | n/a | n/a |
| **GlaS** | | | | | | | | |
| | **PAGET-S** | 0.767 | 0.454 | 0.541 | 0.371 | **0.206** | 0.158 | 0.009 |
| | **PAGET-H** | **0.779** | **0.556** | 0.601 | 0.426 | 0.192 | **0.168** | 0.020 |
| | *Teachers* | 0.719 | 0.498 | 0.407 | 0.234 | 0.033 | 0.000 | 0.045 |
| | Hovernet (pannuke) | 0.687 | 0.505 | 0.597 | n/a | n/a | n/a | n/a |
| | Hovernet (monusac) | 0.525 | n/a | 0.543 | 0.513 | n/a | n/a | **0.084** |
| | HD-Yolo (lung) | 0.372 | 0.143 | 0.208 | 0.204 | n/a | n/a | n/a |
| | HD-Yolo (breast) | 0.638 | 0.409 | **0.613** | **0.526** | n/a | n/a | n/a |
| **CoNSeP** | | | | | | | | |
| | **PAGET-S** | 0.894 | 0.609 | 0.734 | 0.662 | 0.424 | **0.470** | 0.378 |

|  | | | | | | | |
|---|---|---|---|---|---|---|---|
| **PAGET-H** | **0.904** | 0.630 | 0.746 | 0.657 | **0.429** | <u>0.450</u> | 0.367 |
| *Teachers* | 0.710 | <u>0.677</u> | 0.621 | 0.513 | 0.000 | 0.000 | 0.249 |
| Hovernet (pannuke) | 0.860 | **0.736** | **0.831** | n/a | n/a | n/a | n/a |
| Hovernet (monusac) | 0.714 | n/a | <u>0.771</u> | 0.674 | n/a | n/a | **0.386** |
| HD-Yolo (lung) | 0.129 | 0.157 | 0.081 | 0.094 | n/a | n/a | n/a |
| HD-Yolo (breast) | 0.619 | 0.382 | 0.688 | 0.589 | n/a | n/a | n/a |
| **CRAG** | | | | | | | |
| **PAGET-S** | **0.877** | 0.611 | 0.712 | 0.518 | **0.399** | **0.374** | **0.343** |
| **PAGET-H** | <u>0.864</u> | **0.695** | <u>0.737</u> | <u>0.546</u> | <u>0.391</u> | <u>0.366</u> | <u>0.340</u> |
| *Teachers* | 0.772 | <u>0.688</u> | 0.563 | 0.335 | 0.108 | 0.000 | 0.150 |
| Hovernet (pannuke) | 0.794 | 0.663 | **0.808** | n/a | n/a | n/a | n/a |
| Hovernet (monusac) | 0.637 | n/a | 0.728 | **0.619** | n/a | n/a | 0.065 |
| HD-Yolo (lung) | 0.121 | 0.025 | 0.095 | 0.115 | n/a | n/a | n/a |
| HD-Yolo (breast) | 0.375 | 0.257 | 0.383 | 0.315 | n/a | n/a | n/a |

**Table 7 MCC values in the KCCRC dataset.** Bold and italic values indicate the models with the highest and second-highest performance, respectively, for each cell type.

| Model | endothelial cell | leukocyte | lymphocyte | plasma cell | myeloid | eosinophil | neutrophil | fibroblast |
|---|---|---|---|---|---|---|---|---|
| **PAGET-S** | **0.251** | **0.507** | <u>0.539</u> | **0.498** | **0.343** | 0.331 | <u>0.385</u> | <u>0.249</u> |
| **PAGET-H** | 0.226 | <u>0.497</u> | 0.533 | <u>0.484</u> | <u>0.320</u> | 0.331 | **0.388** | 0.213 |
| *Teachers* | <u>0.237</u> | 0.487 | **0.600** | 0.477 | 0.314 | 0.267 | 0.296 | 0.237 |
| Hovernet (pannuke) | n/a | 0.377 | n/a | n/a | n/a | n/a | n/a | **0.257** |
| Hovernet (monusac) | n/a | 0.404 | 0.307 | n/a | n/a | n/a | 0.357 | n/a |
| HD-Yolo (lung) | n/a | 0.378 | 0.364 | n/a | n/a | n/a | n/a | 0.142 |
| HD-Yolo (breast) | n/a | 0.463 | 0.291 | n/a | n/a | n/a | n/a | 0.229 |
| Cerberus | n/a | 0.453 | 0.261 | 0.387 | n/a | **0.343** | n/a | 0.194 |

*Cell counting and time-accuracy tradeoffs*

We next investigated the performance tradeoffs between PAGET-S and PAGET-H in terms of cell counting accuracy and computational efficiency. While PAGET-S offers rapid processing, it lacks the capability for nucleus instance segmentation, precluding direct cell counting. Conversely, PAGET-H provides this functionality but at the cost of increased processing time due to its multi-step approach.

To address this limitation in PAGET-S, we hypothesized that the pixel area of segmented regions or the number of disconnected regions could serve as a proxy for cell count. We tested this hypothesis by examining the correlation between the segmented area in PAGET-S and the cell counts obtained from PAGET-H across various cell types and datasets (Fig. 5a).

Our analysis revealed strong correlations for most cell types, particularly for immune cells ($R^2 > 0.85$ for lymphocytes, plasma cells, myeloid cells, and neutrophils). This high correlation suggests that PAGET-S can provide rapid and reasonably accurate estimates of cell populations, making it a valuable tool for high-throughput quantification of the tumor microenvironment (TME).

We observed slightly lower correlations for fibroblasts and tumor cells. For fibroblasts, this discrepancy may stem from their variable size and elongated morphology, making accurate area-based counting challenging. In the case of tumor cells, the reduced correlation can be attributed to the significant variation in cell size and morphology resulting from genetic alterations, leading to greater dispersion in the area-to-count relationship.

To further elucidate the performance characteristics of both models across different datasets, we analyzed the estimated mean area per cell for various datasets (Fig. 5b). The observed variations across datasets highlight the importance of considering tissue-specific factors such as staining protocols and preparation techniques when interpreting results. These findings suggest that PAGET-H may offer superior performance when dealing with samples from diverse sources or when analyzing cases with atypical cellular morphologies.

In terms of computational efficiency, PAGET-H required an average processing time of 2.1 seconds per image, while PAGET-S completed the analysis in 0.14 seconds, representing a 15-fold speed improvement (Fig.5c). This substantial difference in processing time underscores the importance of selecting the appropriate model based on the specific requirements of the analysis and available computational resources.

Area-based cell counting provides a complementary approach for estimating cell counts from semantic segmentation. While it performs well for cells with consistent sizes like lymphocytes, the connected components approach offers better accuracy for cells with variable morphology such as fibroblasts and epithelial cells. The strong correlations observed within individual datasets, despite differences in their scaling factors, suggest that these relationships could be standardized through dataset-specific calibration. However, this approximation approach may not be suitable for analyses requiring precise nuclear boundaries, such as nuclear shape or texture analysis, where PAGET-H's accurate nucleus segmentation would be necessary.

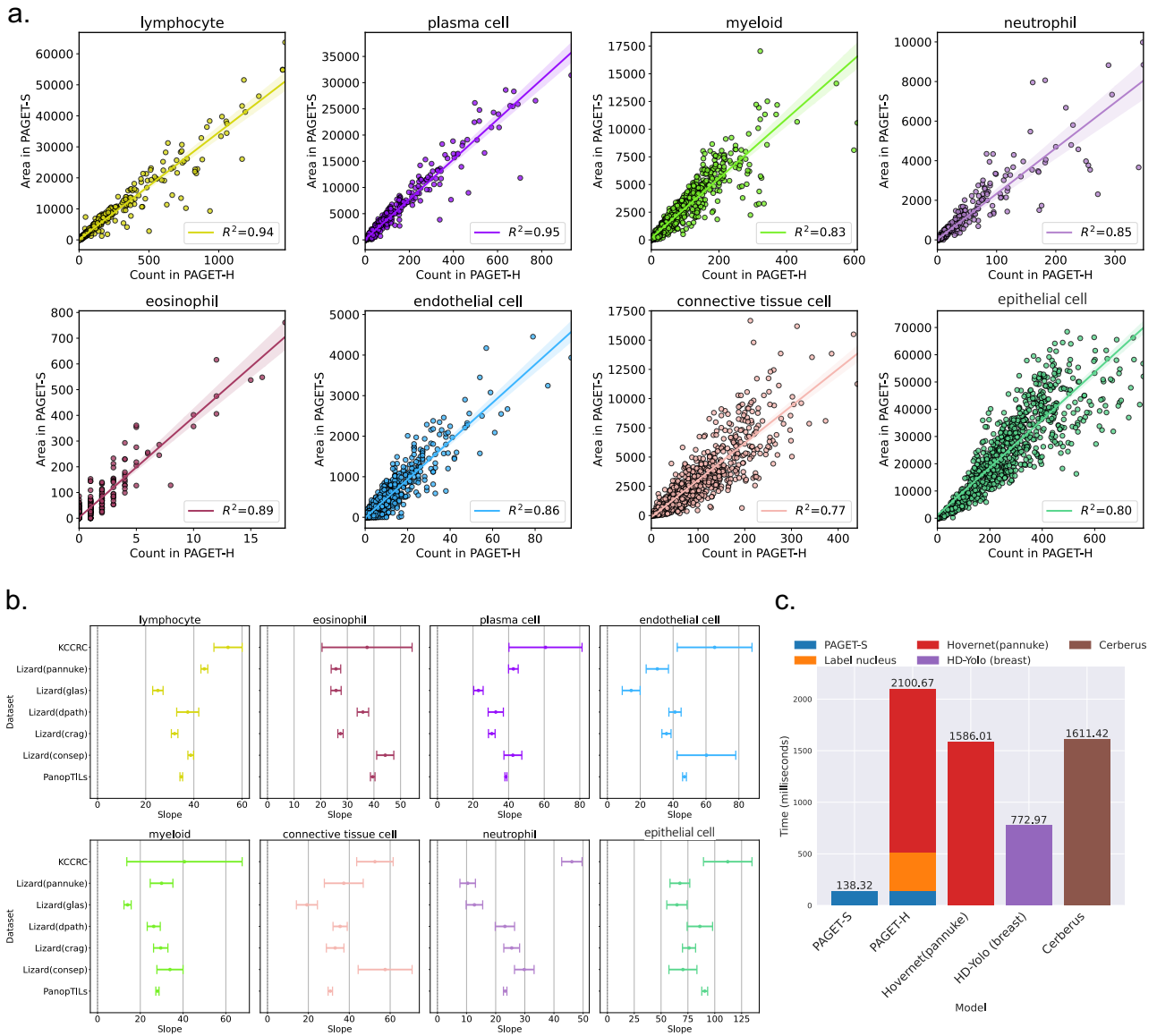

**Fig. 5 Accuracy of area-based cell counting and computation time of PAGET.** a. Correlation between cell area by PAGET-S and cell counts by PAGET-H in various cell types in PanopTILs dataset. b. estimated mean area per cell among different dataset. c. comparison of computation time across models, showing the total processing duration from image input through segmentation inference to image saving.

*Quantitative evaluation of tumor microenvironment using PAGET-S*

Finally, we applied PAGET-S to WSIs of uterine corpus endometrial carcinoma (UCEC) from the Cancer Genome Atlas (TCGA). Previous studies have demonstrated that various somatic mutations in tumor cells can significantly influence both the composition and spatial distribution of the tumor microenvironment[25]. Understanding these mutation-driven effects on the TME is crucial for elucidating cancer progression mechanisms and developing targeted therapies, as altered tumor-stromal interactions could lead to the emergence of novel therapeutic targets and drug resistance mechanisms. To systematically investigate these relationships and demonstrate the utility of PAGET-S, we conducted a comprehensive analysis examining the associations between TME metrics derived from PAGET-S outputs and known somatic driver mutations (Fig. 6).

The analysis confirmed several previously reported relationships. For example, we found that lymphocytes in both the tumor area and tumor margin were significantly enriched in cases with POLE and MSH6 or PMS2 mutations. These alterations lead to an ultra- or hyper-mutated phenotype, generating abundant neoantigens that enhance immune recognition and response, which explains the elevated peri-tumoral and intra-epithelial lymphocytic infiltration observed in POLE-mutated and mismatch repair deficient endometrial cancers. Additionally, we found that mitotic cells in the tumor area were significantly enriched in cases with TP53 mutations, consistent with previous findings that p53 abnormal endometrial cancers are characterized by high mitotic activity due to loss of cell cycle control[26]. A representative histological image of a case with TP53 mutation is shown in Fig. 7.

We demonstrated an example of the case with CTNNB1 mutation in Supplementary Fig. 2. It is known that activation of Wnt/β-catenin pathway in tumor leads to noninflammatory milieu and β-catenin levels and CD8 T cell infiltration have inverse correlations[27]. Shukla et al. reported that CTNNB1 mutations were enriched in CN-low/endometrioid tumors with lower neoantigen load[28]. Our findings of reduced lymphocyte and total leukocyte counts in CTNNB1-mutated cases compared to wild-type (Supplementary Fig.3) are consistent with these reports (Supplementary Fig.4).

We also found that ARID1A mutations were associated with increased lymphocyte infiltration. ARID1A, a key component of the SWI/SNF chromatin-remodeling complex, activates mismatch repair (MMR) by enlisting MSH2[29], and thus ARID1A deletion could lead to MMR deficiency and MSI in several types of cancer, including endometrioid carcinoma of the uterus[30]. Additionally, increased PD-L1 expression in ARID1A deficient cells is attributed to upregulation of double-strand breaks (DSBs)[30]. These mechanisms result in enhanced tumor mutability and T-lymphocyte infiltration[28–30], which is consistent with our finding. In addition to the elevated lymphocyte infiltration, of particular interest was the observation of a remarkable increase in neutrophils and myeloid cells, which represented one of the most distinctive immune cell patterns among all gene mutations analyzed. As shown in Fig. 8, infiltration of neutrophils and myeloid cells was observed in the tumor area and peripheral area. However, some cells that have spilled into the glandular lumen were also counted, so further validation will be conducted in the future.

The ratio of "endothelial cells" in the peripheral area increased with FGFR2 mutations (Supplementary Fig. 5). FGFR2 had been shown to be activated in a number of cancers due to gene amplification and point mutation and Byron et al. reported somatic activating FGFR2 mutation in 16% of endometrioid endometrial cancers[31]. FGFR2 are not constitutively active in non-malignant cells[32]. The oncogenic role of FGF-FGFR signaling in driving cell proliferation, survival, migration and invasion is mediated by the upregulation of FGF, FGFR genetic alterations, angiogenesis and immune evasion in the tumor microenvironment[33,34]. In recent years, a variety of angiogenesis inhibitors targeting factors such as FGF and FGFR had been developed[33]. When testing the therapeutic efficacy of these agents, evaluating the endothelial cells surrounding the tumor was expected to provide valuable information, such as identifying which types of cases might benefit from the therapy.

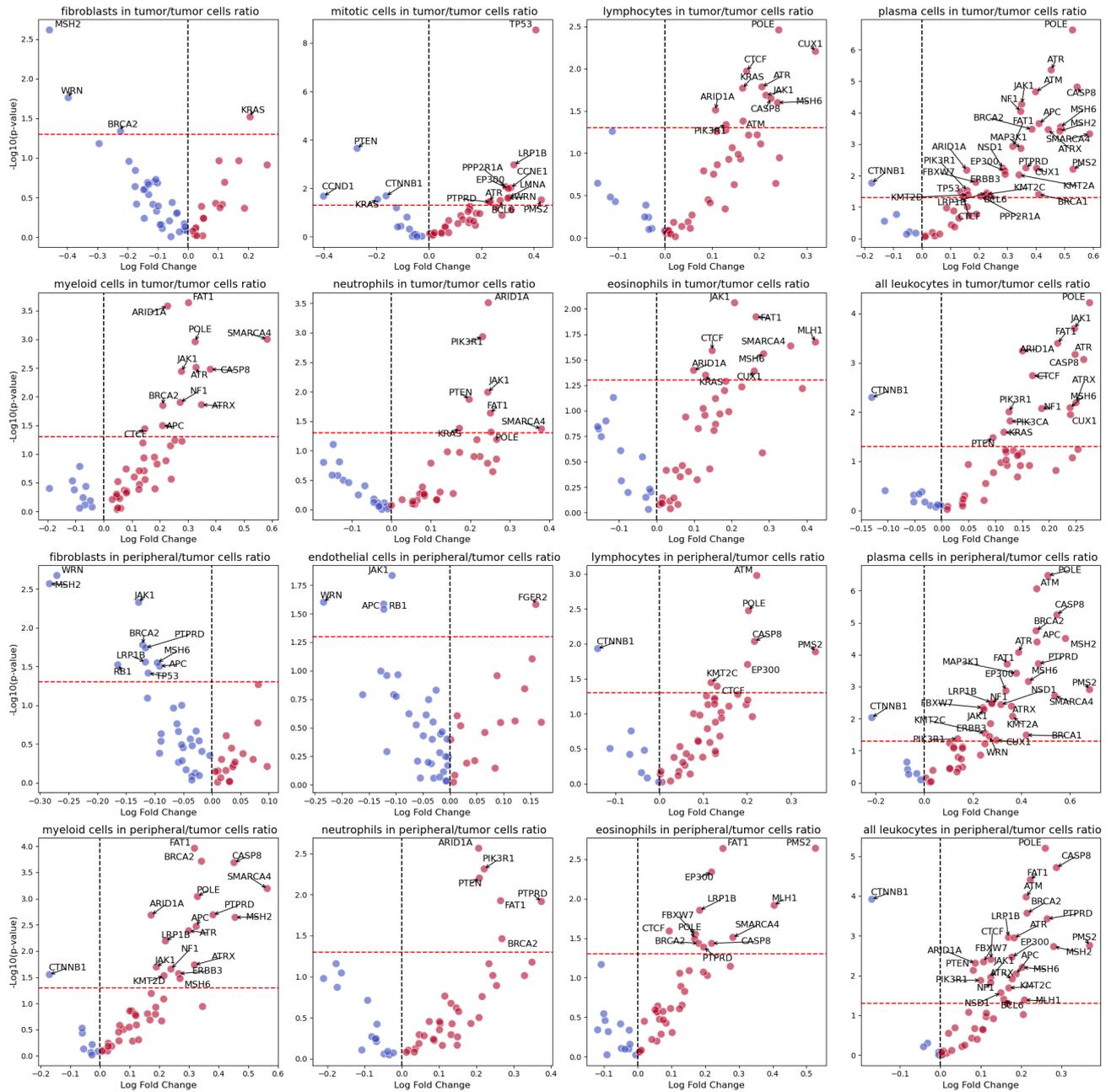

**Fig. 6. The association between tumor microenvironment metric and driver mutations in 445 UCEC cases.** Here, "in tumor" represents the ratio of the counts of each cell type to tumor cell count within the whole tumor tissue, while "in peripheral" denotes the ratio of cell density to tumor cell count within the 50 μm surrounding tumor area. The nominal p-values were calculated using the Mann-Whitney U statistical test for each driver genes.

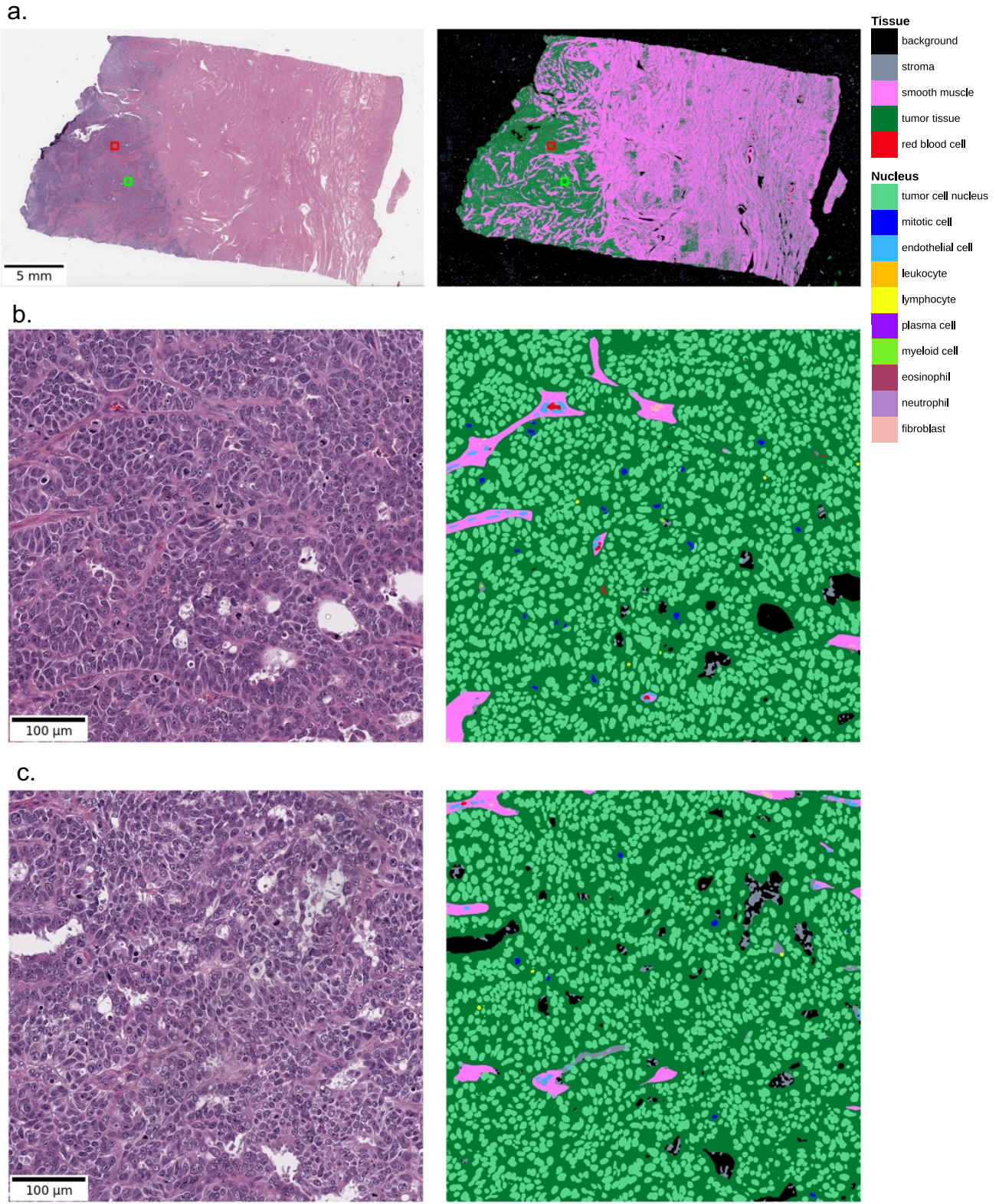

**Fig. 7. Example of the case with *TP53* mutation.** a. Whole slide image thumbnail (TCGA-E6-A8L9-01Z-00-DX1.svs). b. region highlighted by red box in a. c. region highlighted by green box in a. Both b. and c. showed presence of mitotic cells (blue) in the tumor area.

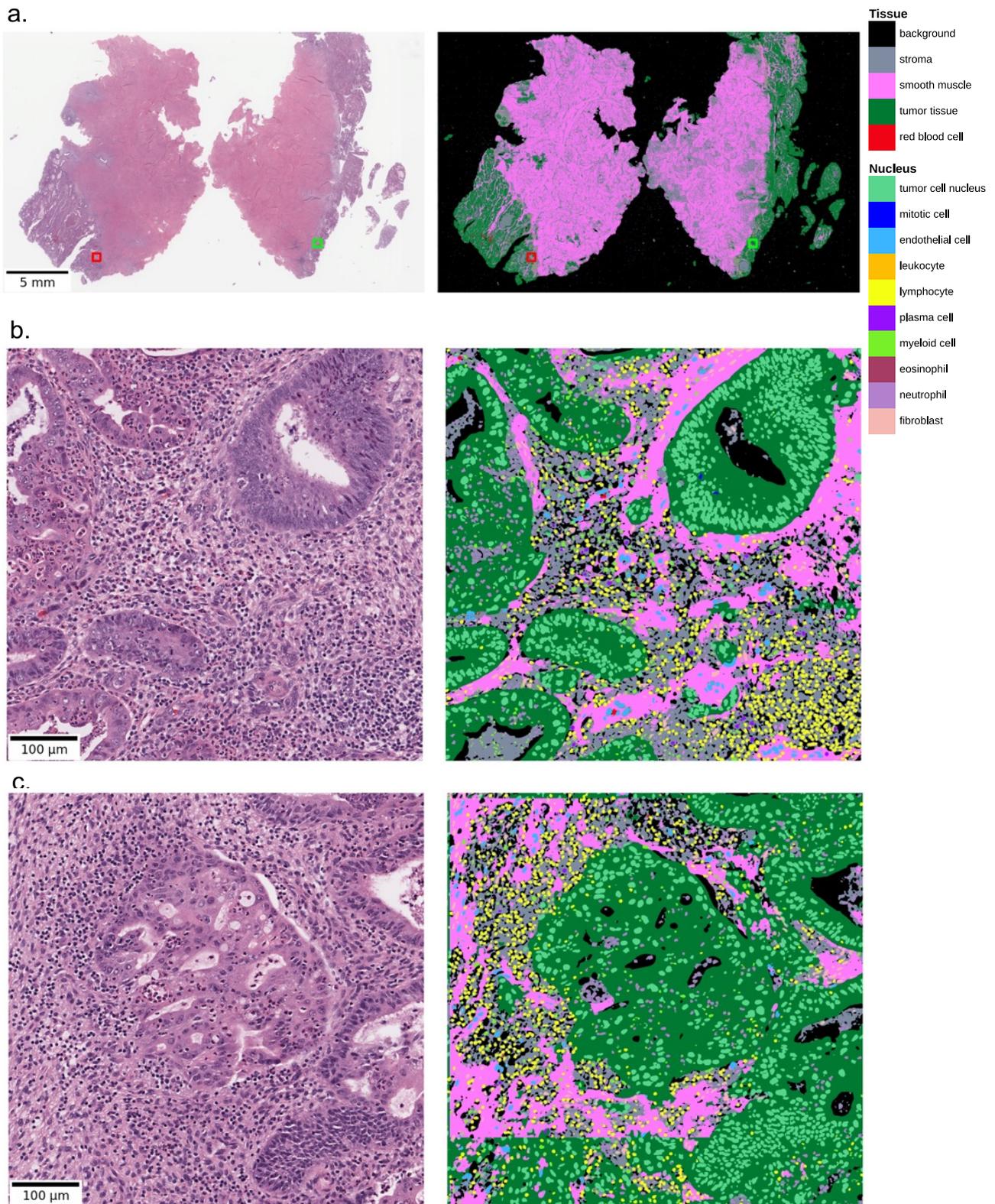

**Fig. 8. Example of the case with *ARID1A* mutation.** a. Whole slide image thumbnail (TCGA-D1-A17Q-01Z-00-DX1.svs). b. region highlighted by red box in a. c. region highlighted by green box in a. Both b. and c. showed infiltration of not only lymphocytes and neutrophils but also myeloid cells.

**Conclusion**

In this study, we introduced PAGET, a novel approach for comprehensive analysis of tumor microenvironment (TME) analysis in histopathology. PAGET's cell hierarchy-aware aggregated distillation methodology enables simultaneous identification and classification of multiple tissue and cell types in H&E-stained slides, addressing the longstanding challenge of comprehensive, multi-class segmentation in histopathological analysis. The approach's two variants, PAGET-S for semantic segmentation and PAGET-H for panoptic segmentation, offer flexibility in balancing processing speed and information granularity. Extensive evaluation on external datasets demonstrated PAGET's superior performance across various cell types and its ability to accurately capture tumor tissue distribution and immune cell infiltration patterns at scale.

PAGET's detailed analysis of the tumor microenvironment (TME) may contribute to our understanding of cancer biology. By offering a more comprehensive analysis of the TME while addressing some current limitations in digital pathology, PAGET represents a step forward in this field. This approach has the potential to aid in identifying prognostic factors in cancer and may help predict responsiveness to various therapies. Future studies investigating correlations between PAGET's analytical results and clinical outcomes could establish it as a valuable tool in cancer research.

*Ethics approval*

This study was conducted in accordance with the Declaration of Helsinki and approved by the Institutional Review Boards (IRBs) of The University of Tokyo (approval number: 2381 and 2019158NI), Japanese Red Cross Medical Center (approval number: 1414), and Kanagawa Cancer Center (approval number: 2020-118).

An opt-out approach was adopted for this study. Information about the study and the option to opt out was provided to patients through public notices in the hospitals and on their websites. Patients were given the opportunity to refuse the use of their clinical data and pathological specimens for research purposes. Those who did not opt out were considered to have given implicit consent for their data to be used in this study.

The ethics committees approved this opt-out procedure for this retrospective analysis of anonymized data and images. All procedures were performed in compliance with relevant laws and institutional guidelines.

**Acknowledgement**

This work was supported by the AMED Practical Research for Innovative Cancer Control grant number JP 24ck0106873 and JP 24ck0106904 to S.I., and JSPS KAKENHI Grant-in-Aid for Scientific Research (B) grant number 21H03836 to D.K.

**Supplementary Table. 1 Class correspondence in evaluation datasets.**

| Dataset | class in eval | class in dataset | PAGET | HoverNet (pannuke) | HoverNet(monusac) | HD-Yolo | Cerberus |
|---|---|---|---|---|---|---|---|
| PanopTILs | epithelial tissue | cancerous epithelium, normal epithelium, cancer nucleus, normal epithelial nucleus | epi, epi_n | N/A | N/A | N/A | N/A |
| | epithelial cell | cancer nucleus, normal epithelial nucleus | epi_n | neopla, no-neo | epi | tumor | epithelial |
| | connective tissue cell | stromal nucleus, large stromal nucleus | endo, fib | connec | N/A | stromal | connective tissue cell |
| | leukocyte | lymphocyte nucleus, plasma cell / large TIL nucleus | lym, pls, mye, eos, neu | inflam | lym, macro, neut | sTILs, macrophage | neutrophil, lymphocyte, plasma cell, eosinophil |
| | lymphocyte | lymphocyte nucleus | lym | N/A | lym | sTILs | lymphocyte |
| | plasma cell | plasma cell / large TIL nucleus | pls | N/A | N/A | N/A | plasma cell |
| Lizard | epithelial cell | epithelial | epi_n | neopla, no-neo | epi | tumor | N/A (data leak) |
| | connective tissue cell | connective | endo, fib | connec | N/A | stromal | N/A (data leak) |
| | leukocyte | lymphocyte, plasma, neutrophil, eosinophil | lym, pls, mye, eos, neu | inflam | lym, macro, neut | sTILs, macrophage | N/A (data leak) |
| | lymphocyte | lymphocyte | lym | N/A | lym | sTILs | N/A (data leak) |
| | plasma cell | plasma | pls | N/A | N/A | N/A | N/A (data leak) |
| | eosinophil | eosinophil | eos | N/A | N/A | N/A | N/A (data leak) |
| | neutrophil | neutrophil | neu | N/A | neutrophil | N/A | N/A (data leak) |
| KCCRC | endothelial cell | endothelial cell | endo | N/A | N/A | N/A | N/A |
| | leukocyte | lymphocyte, plasma cell, myeloid cell, eosinophil, neutrophil | lym, pls, mye, eos, neu | inflam | lym, macro, neut | sTILs, macrophage | neutrophil, lymphocyte, plasma cell, eosinophil |
| | lymphocyte | lymphocyte | lym | N/A | lym | sTILs | lymphocyte |
| | plasma cell | plasma cell | pls | N/A | N/A | N/A | plasma cell |
| | myeloid cell | myeloid cell, eosinophil, neutrophil | mye | N/A | N/A | N/A | N/A |
| | eosinophil | eosinophil | eos | N/A | N/A | N/A | eosinophil |
| | neutrophil | neutrophil | neu | N/A | neutrophil | N/A | neutrophil |
| | mitotic cell | mitotic cell | mit | N/A | N/A | N/A | N/A |

**Supplementary Table. 2 Analyzed driver mutations in UCEC cases**

| driver gene | cases(n) | frequency(%) |
|---|---|---|
| PTEN | 305 | 68.5 |
| PIK3CA | 231 | 51.9 |
| ARID1A | 185 | 41.6 |
| TP53 | 156 | 35.1 |
| PIK3R1 | 135 | 30.3 |
| CTCF | 95 | 21.3 |
| CTNNB1 | 94 | 21.1 |
| KRAS | 90 | 20.2 |
| FBXW7 | 75 | 16.9 |
| ZFHX3 | 74 | 16.6 |
| KMT2D | 62 | 13.9 |
| FGFR2 | 55 | 12.4 |
| JAK1 | 52 | 11.7 |
| ATM | 52 | 11.7 |
| LRP1B | 51 | 11.5 |
| FAT1 | 50 | 11.2 |
| POLE | 47 | 10.6 |
| PPP2R1A | 46 | 10.3 |
| NF1 | 46 | 10.3 |
| KMT2C | 43 | 9.7 |
| NSD1 | 41 | 9.2 |
| APC | 37 | 8.3 |
| CCNE1 | 36 | 8.1 |
| BRCA2 | 35 | 7.9 |
| MYC | 35 | 7.9 |
| BCOR | 35 | 7.9 |
| ERBB2 | 35 | 7.9 |
| EP300 | 34 | 7.6 |
| MAP3K1 | 33 | 7.5 |
| CASP8 | 31 | 7 |
| ATRX | 30 | 6.7 |
| RB1 | 29 | 6.5 |
| PTPRD | 29 | 6.5 |
| CCND1 | 29 | 6.5 |
| WRN | 28 | 6.3 |
| BCORL1 | 28 | 6.3 |
| MSH6 | 27 | 6.1 |
| LMNA | 27 | 6.1 |
| *SPOP* | 26 | 5.8 |
| *ERBB3* | 26 | 5.8 |
| *BCL6* | 26 | 5.8 |
| *DICER1* | 25 | 5.6 |
| *CUX1* | 24 | 5.4 |
| *ATR* | 23 | 5.2 |
| *CREBBP* | 23 | 5.2 |
| *KMT2A* | 23 | 5.2 |
| *MSH2* | 21 | 4.7 |
| *SMARCA4* | 13 | 2.9 |
| *BRCA1* | 11 | 2.5 |
| *PMS2* | 11 | 2.5 |
| *MLH1* | 11 | 2.5 |

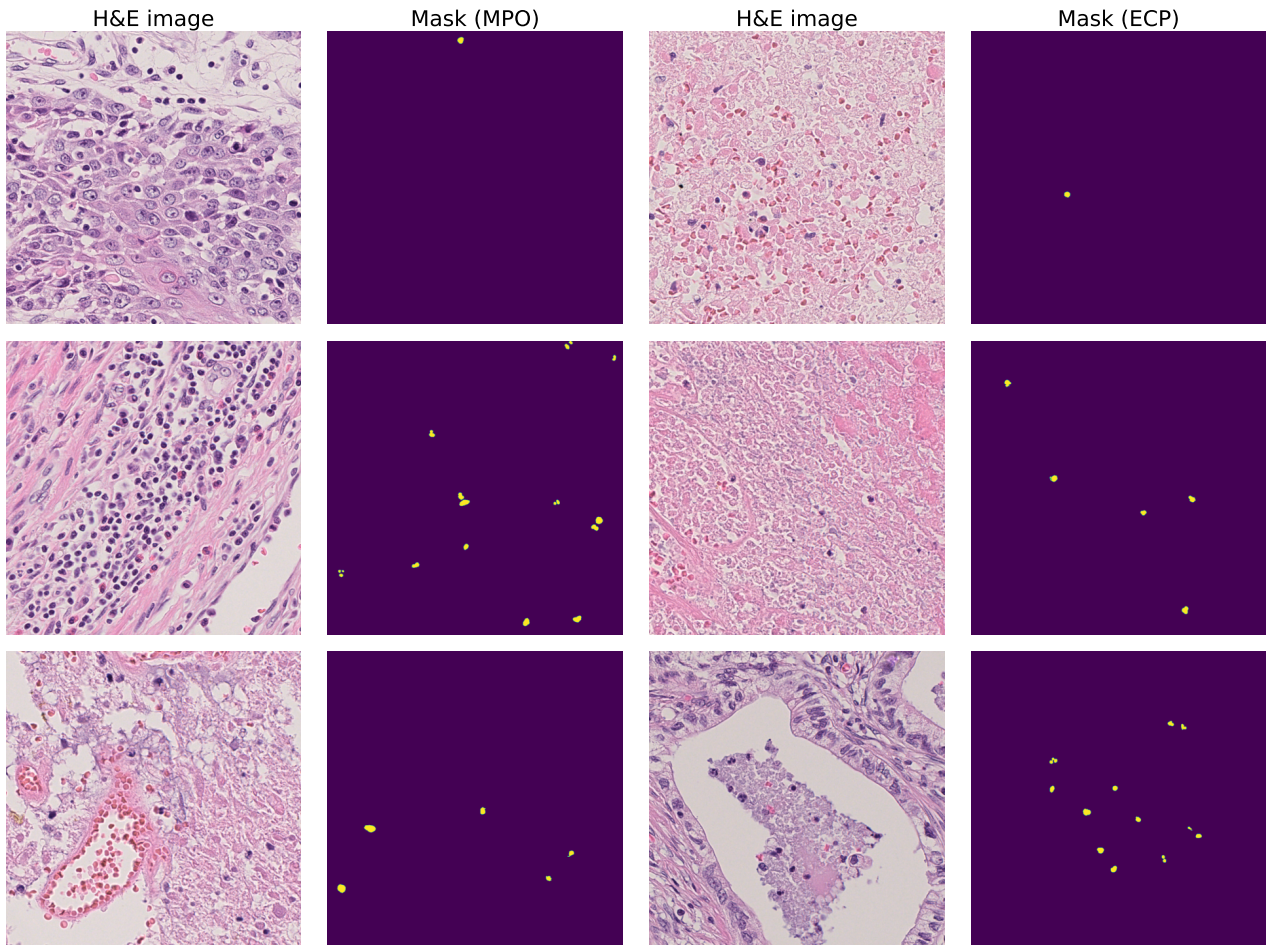

**Supplementary Fig. 1 Samples of MPO and ECP masks.**

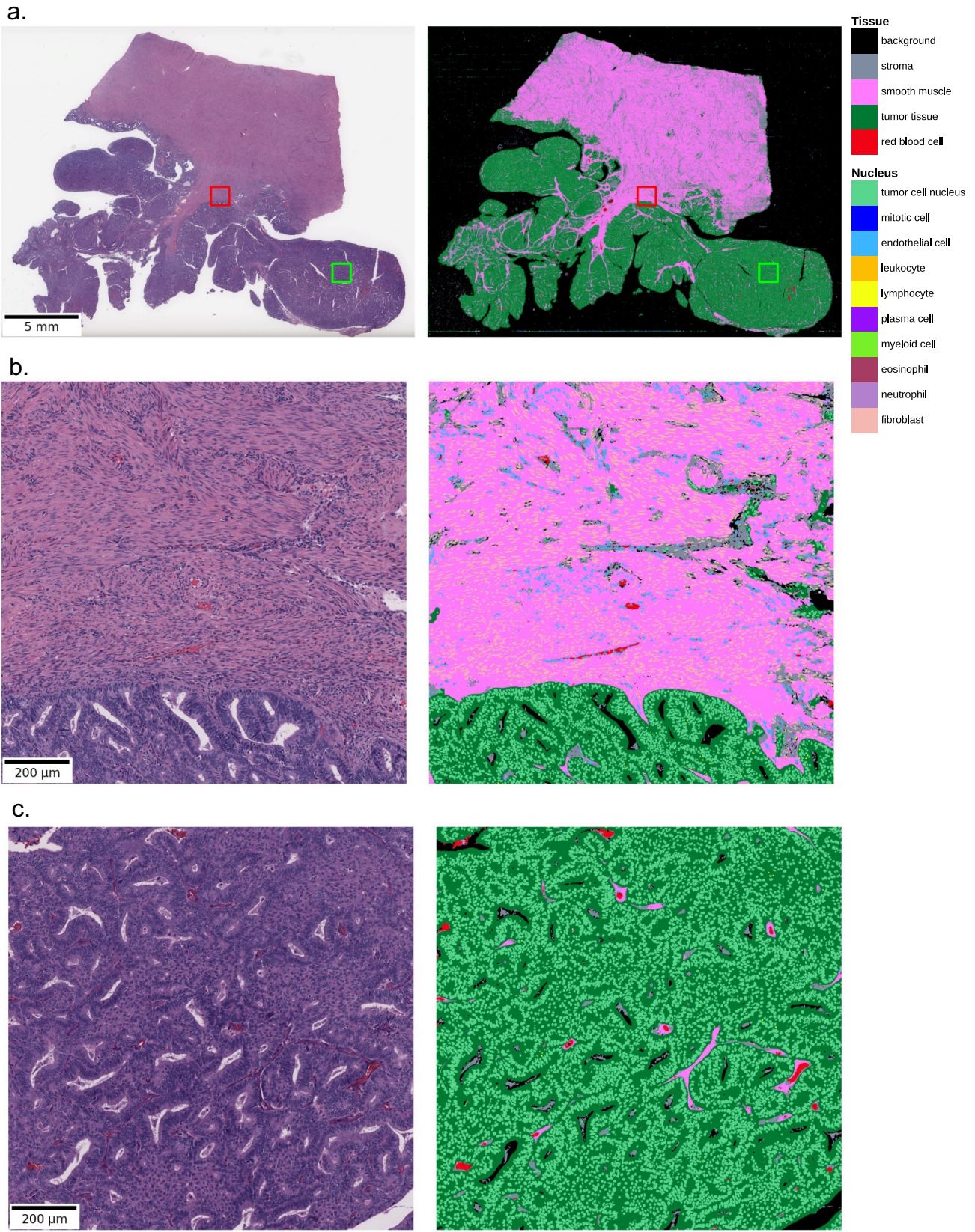

**Supplementary Fig. 2. Example of the case with *CTNNB1* mutation.** a. Whole slide image thumbnail (TCGA-BG-A0LW-01Z-00-DX1.svs). b. region highlighted by red box in a. c. region highlighted by green box in a. Infiltration of immune cells other than fibroblasts and endothelial cells was infrequent in the peripheral area (b) and tumor area (c).

a.

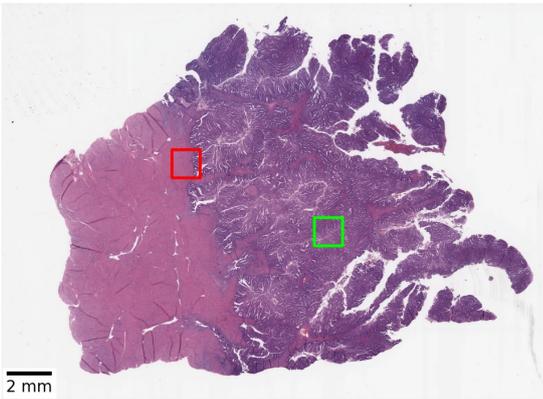
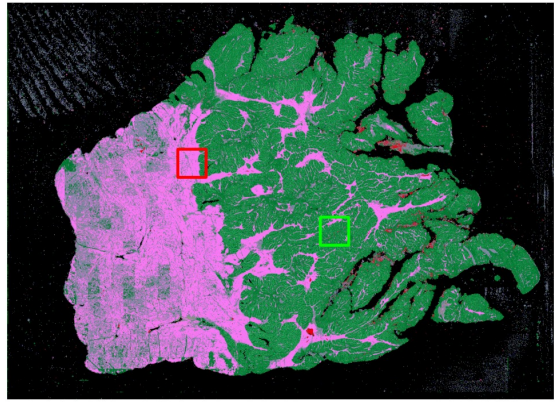

b.

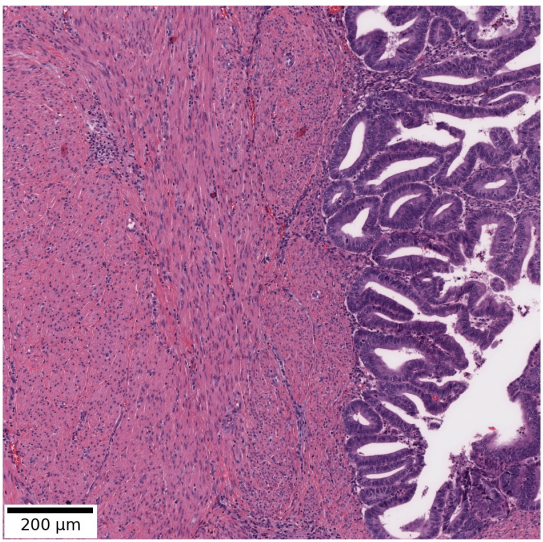
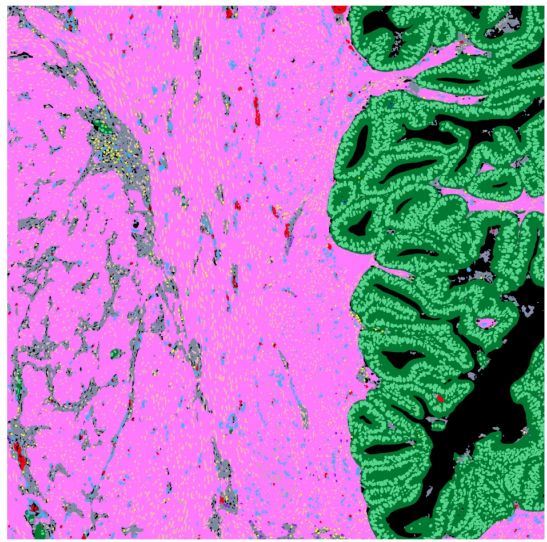

c.

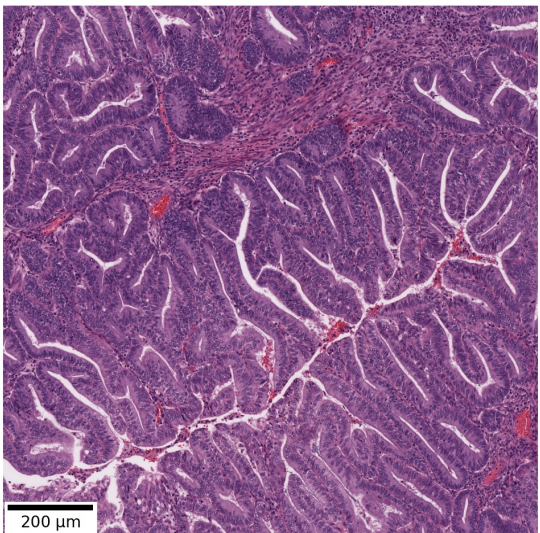
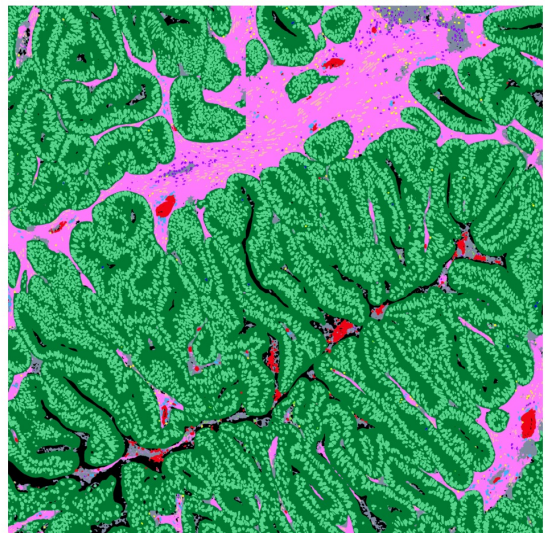

**Supplementary Fig.3. Example of the case without *CTNNB1* mutation.** a. Whole slide image thumbnail (TCGA-DI-A1NN-01Z-00-DX1.svs). b. region highlighted by red box in a. c. region highlighted by green box in a. Scattered lymphocytes, myeloid cells, and plasma cells can be observed in the peripheral area (b) and tumor area (c).

a.

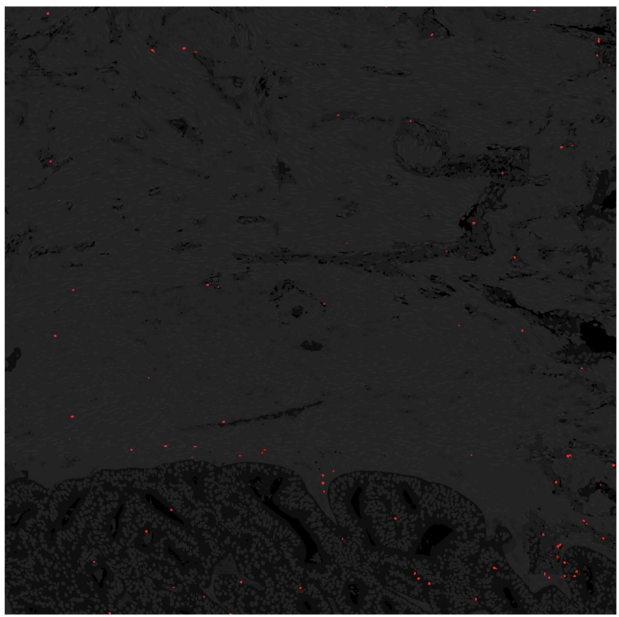 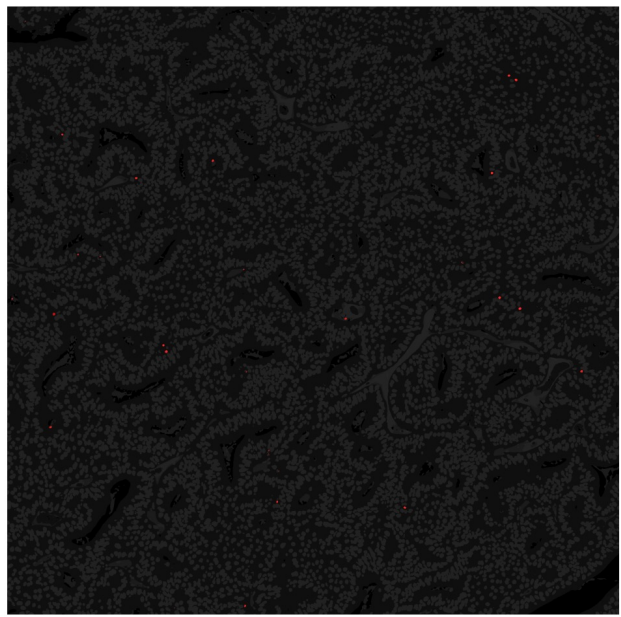

b.

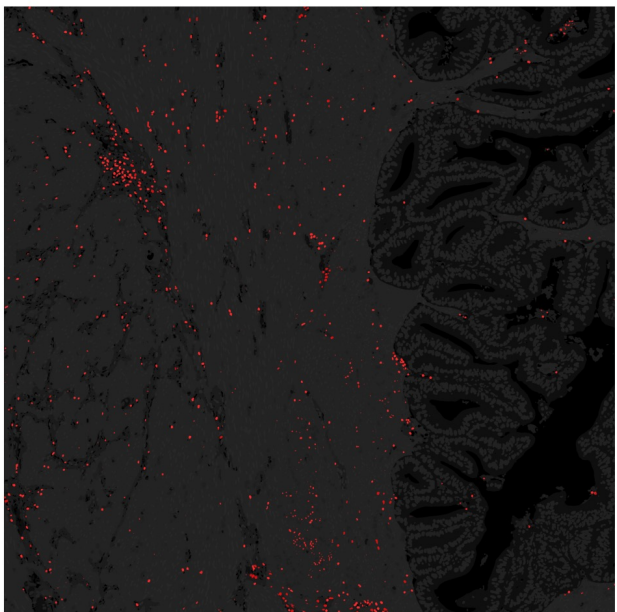 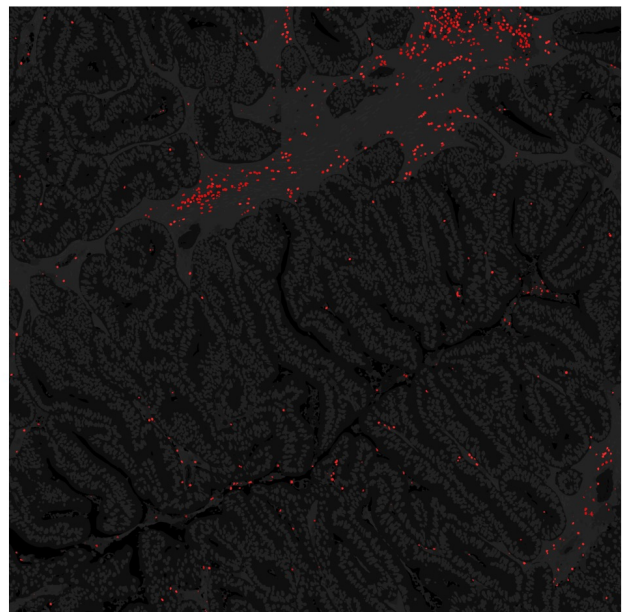

**Supplementary Fig.4. Comparison of immune cell distribution between CTNNB1-mutated and wild-type cases.** a. Grayscale segmentation results by PAGET from CTNNB1-mutated cases (corresponding to Supplementary Fig.2 b and c). b. Grayscale segmentation results by PAGET from CTNNB1 wild-type cases (corresponding to Supplementary Fig.3 b and c). Red dots indicate immune cells (lymphocytes, myeloid cells, and plasma cells).

a.

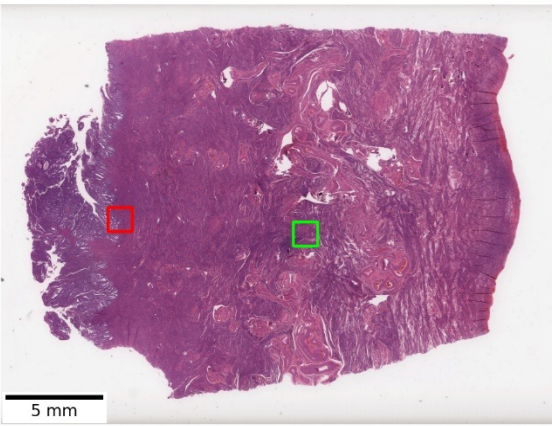 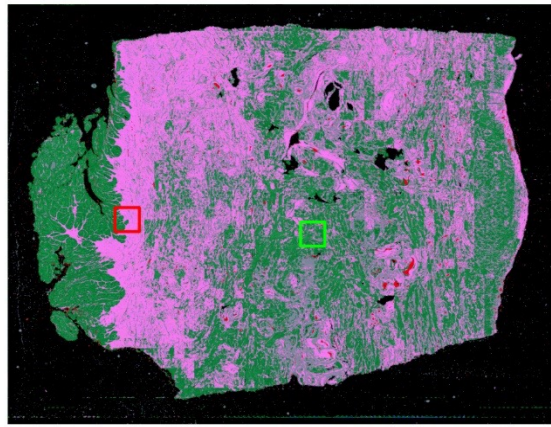

b.

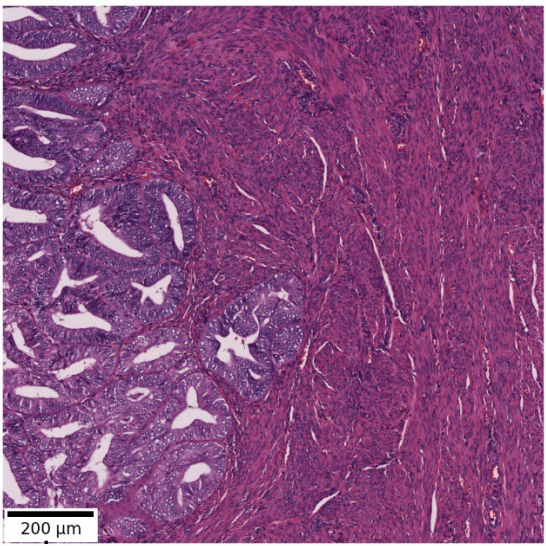 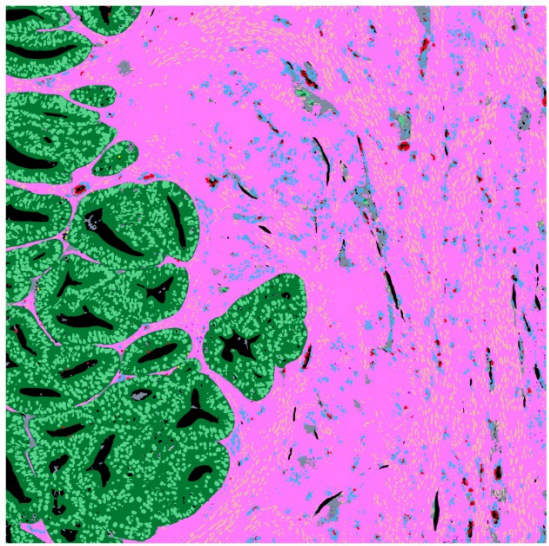

c.

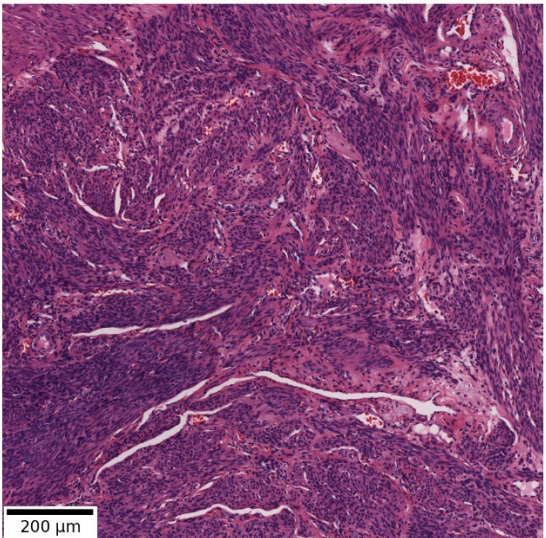 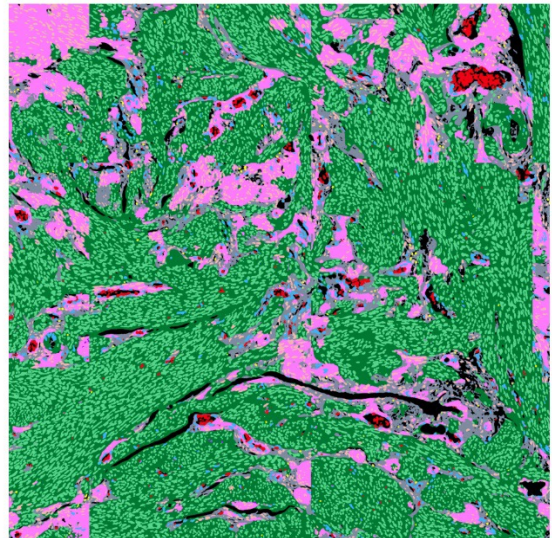

**Supplementary Fig. 5. Example of the case with *FGFR2* mutation.** a. Whole slide image thumbnail (TCGA-A5-A0GU-01Z-00-DX1.svs). b. region highlighted by red box in a. c. region highlighted by green box in a. Both b. and c. showed prominent endothelial cell proliferation in the peripheral area.